\documentclass[10pt,twocolumn,letterpaper]{article}

\usepackage{cvpr}              

\usepackage{booktabs}
\usepackage{graphicx}
\usepackage{array}
\usepackage{makecell}
\usepackage{amssymb}
\usepackage{enumitem}

\definecolor{cvprblue}{rgb}{0.21,0.49,0.74}
\usepackage[pagebackref,breaklinks,colorlinks,allcolors=cvprblue]{hyperref}

\definecolor{negative}{HTML}{E52B50}
\definecolor{positive}{HTML}{6AB2A4}

\def\paperID{13} 
\def\confName{CVPR}
\def\confYear{2025}

\title{Towards Faster and More Compact Foundation Models\\for Molecular Property Prediction}

\author{Yasir Ghunaim, Andrés Villa, Gergo Ignacz, Gyorgy Szekely, Motasem Alfarra, and
Bernard Ghanem\\
King Abdullah University of Science and Technology (KAUST)
}

\newcommand\blfootnote[1]{%
  \begingroup
  \renewcommand\thefootnote{}\footnote{#1}%
  \addtocounter{footnote}{-1}%
  \endgroup
  }

\begin{document}
\maketitle

\begin{abstract}
Advancements in machine learning for molecular property prediction have improved accuracy but at the expense of higher computational cost and longer training times. Recently, the Joint Multi-domain Pre-training (JMP) foundation model has demonstrated strong performance across various downstream tasks with reduced training time over previous models. Despite JMP's advantages, fine-tuning it on molecular datasets ranging from small-scale to large-scale requires considerable time and computational resources. In this work, we investigate strategies to enhance efficiency by reducing model size while preserving performance. To better understand the model's efficiency, we analyze the layer contributions of JMP and find that later interaction blocks provide diminishing returns, suggesting an opportunity for model compression. We explore block reduction strategies by pruning the pre-trained model and evaluating its impact on efficiency and accuracy during fine-tuning. Our analysis reveals that removing two interaction blocks results in a minimal performance drop, reducing the model size by 32\% while increasing inference throughput by 1.3×. These results suggest that JMP-L is over-parameterized and that a smaller, more efficient variant can achieve comparable performance with lower computational cost. Our study provides insights for developing lighter, faster, and more scalable foundation models for molecular and materials discovery. The code is publicly available at: \href{https://github.com/Yasir-Ghunaim/efficient-jmp}{github.com/Yasir-Ghunaim/efficient-jmp}. \blfootnote{Correspondence to: \texttt{yasir.ghunaim@kaust.edu.sa}}
\end{abstract}

\section{Introduction}

Molecular property prediction using density functional theory (DFT) and molecular dynamics (MD) calculations plays a crucial role in the discovery of novel materials, including pharmaceutical drugs~\cite{Sabe2021Current}, catalysts~\cite{Nrskov2011Density, tran2023open, chanussot2021open}, metal-organic frameworks~\cite{Rosen2022}, and polymers~\cite{Sharma2014}. However, the high computational cost of DFT and MD calculations limits their feasibility for large-scale, high-throughput searches. To address this challenge, machine learning potentials have been developed to accelerate DFT and MD calculations~\cite{Behler2007, Bogojeski2020}, leveraging the latest large-scale datasets \cite{kolluru2022openchallengesdevelopinggeneralizable}, such as Open Catalyst 2020 (OC20)~\cite{chanussot2021open}, Open Catalyst 2022 (OC22)~\cite{tran2023open}, and ODAC23~\cite{sriram2023opendac2023dataset}. However, training models from scratch for different tasks remains a major bottleneck for their widespread adoption. Differences in applied DFT theories, molecular system sizes, and chemical diversity increase the molecular system complexity, thus hindering the generalizability and scalability of machine learning models in chemistry.

Recent progress in efficient pre-training strategies~\cite{zaidi2022pretrainingdenoisingmolecularproperty, zhou2023unimol}, the availability of extensive DFT and MD datasets~\cite{tran2023open, chanussot2021open, eastman2023spice, smith2020ani}, and the introduction of specialized chemical benchmarks~\cite{Schreiner2022, dunn2020benchmarking} have led to the emergence of foundation models for molecular property prediction. Foundation models such as the Joint Multi-domain Pre-training (JMP) model~\cite{shoghi2023molecules} and MACE-MP-0 \cite{batatia2023foundation, Batatia2022mace, Batatia2022Design} have demonstrated outstanding performance in diverse molecular tasks. In particular, JMP adapts the pre-train-then-finetune paradigm from vision and language tasks to molecular property prediction. By pre-training on large datasets, the JMP model captures generalizable molecular representations that enable faster fine-tuning for downstream tasks, overcoming the need to train models from scratch for each new application.

Although the large variant of JMP (JMP-L) has outperformed state-of-the-art models on 34 out of 40 tasks, its efficiency in fine-tuning and inference has yet to be addressed. With 160M parameters, JMP-L achieves similar performance to MACE~\cite{Batatia2022mace}, which uses only 3M parameters, suggesting potential over-parameterization. This over-parameterization increases memory and compute requirements and leads to higher carbon emissions~\cite{shoghi2023molecules}, reducing overall sustainability. Although 160M parameters are relatively small compared to vision and language models, the parameter-to-data ratio in molecular ML remains disproportionately large. For instance, MD17 contains only 1,000 training samples with an average of 13 nodes per graph~\cite{shoghi2023molecules}, making a model of this size inefficient for small datasets.

To address these limitations, we perform an in-depth analysis of the efficiency of JMP-L. By examining its interaction block hierarchy, we find that higher-order blocks contribute less to overall performance. This observation aligns with recent findings in large language models, where deeper layers often yield diminishing returns~\cite{gromov2024unreasonableineffectivenessdeeperlayers}. This motivates our exploration of block reduction, a pruning strategy that removes the least important layers to improve efficiency while maintaining accuracy. Additionally, we investigate knowledge distillation techniques tailored to molecular graph neural networks, integrating them with block reduction to assess their combined impact. Although pruning~\cite{liu2022comprehensivegraphgradualpruning} and distillation~\cite{zeng2023molkddistillingcrossmodalknowledge} are widely used in other domains, their application to molecular property prediction, particularly within the pre-train-then-finetune paradigm, remains underexplored.

Our findings reveal that pruning and distillation improve JMP-L's efficiency while preserving comparable performance for most tasks. Specifically, we show that a pruned and distilled variant of JMP-L achieves comparable accuracy to the original model across in-distribution and out-of-distribution downstream tasks. By removing two interaction blocks, we reduce the model size by 32\%, decreasing the parameter count from 160M to 108M, while improving inference throughput by 1.3× compared to the baseline model. These results confirm that JMP-L is over-parameterized for many tasks, and smaller, more efficient versions can achieve similar performance with reduced computational cost. In summary, our contributions are three-fold:
\begin{itemize}
    \item We reduce the number of JMP-L parameters by 32\% to 108M and achieve 1.3× faster inference while maintaining performance.
    \item We evaluate the impact of block reduction and knowledge distillation on pre-training across in-distribution and out-of-distribution downstream tasks.
    \item We demonstrate that later interaction blocks of JMP-L contribute less to performance, supporting the case for model compression. 
\end{itemize}

\section{Related Work}

\subsection{Foundation Models in Molecular Property Prediction}
Pre-trained models have made considerable advancements in developing robust architectures across various domains. Notable examples in the vision domain include ResNet \cite{resnet} and ViT \cite{vit}, which leverage large-scale datasets such as ImageNet \cite{imagenet} to enhance image processing. In contrast, deep learning models for molecular property prediction have primarily been task-specific \cite{batatia2023foundation, maceoff}, limiting their utility as general-purpose pre-trained models. Recently, JMP \cite{shoghi2023molecules} introduced a supervised pre-training strategy on large datasets, establishing a shared knowledge base for various downstream tasks. Built on GemNet-OC~\cite{gasteiger2022gemnet}, JMP is the first large-scale foundation model for molecular property prediction. However, its fine-tuning efficiency remains a challenge, as it requires more than 275 GPU hours to converge \cite{shoghi2023molecules}. In this work, we provide a comprehensive analysis of JMP and propose a more efficient approach to reduce its computational demands, enhancing its accessibility and scalability for broader applications.

\subsection{Efficient Training}

\subsubsection{Pruning} Pruning is a technique used to reduce the size and complexity of a neural network by eliminating weights, neurons, layers, or filters without compromising accuracy \cite{Sietsma1988Neural, cheng2024surveydeepneuralnetwork, blalock2020state}. It is particularly effective when a model is over-parameterized for its task \cite{Sietsma1988Neural}. Structured pruning, which removes entire layers or filters, has been shown to improve memory and computational efficiency in various architectures, including large language models \cite{zhang2024loraprunestructuredpruningmeets, sun2024simpleeffectivepruningapproach}, vision transformers \cite{yu2022width}, and graph neural networks (GNNs) \cite{liu2022comprehensivegraphgradualpruning}. JMP-L, which is based on the GemNet-OC architecture \cite{gasteiger2022gemnet}, consists of an embedding layer, six interaction layers, and three MLP layers. Drawing inspiration from pruning techniques in other domains, we investigate the impact of removing GemNet-OC interaction layers to accelerate fine-tuning and inference while maintaining model performance.

\subsubsection{Knowledge distillation} Knowledge distillation (KD) is a model compression technique that transfers knowledge from a larger teacher model to a smaller student model, aiming to achieve similar performance with reduced computational costs \cite{hinton2015distilling, Bucilua2006Model}. Initially introduced by Bucilua \textit{et al.} \cite{Bucilua2006Model} and later popularized by Hinton \textit{et al.} \cite{hinton2015distilling}, KD has been widely applied in the language \cite{xu2024surveyknowledgedistillationlarge}, vision \cite{habib2024knowledgedistillationvisiontransformers}, and general graph domains \cite{tian2023knowledgedistillationgraphssurvey}, mainly in classification tasks. However, applying KD to large-scale regression tasks such as DFT and MD simulations presents unique challenges~\cite{ekstrom2024accelerating}. Molecular GNNs operate on structured graph data with features distributed across nodes and edges, making direct knowledge transfer from teacher to student challenging~\cite{ekstrom2024accelerating}. These challenges are amplified when the teacher and student models differ considerably in architecture, making feature alignment more difficult. To address these challenges, Ekström \textit{et al.}~\cite{ekstrom2024accelerating} propose specialized loss functions—\textit{node2node (n2n)}, \textit{edge2edge (e2e)}, \textit{edge2node (e2n)}, and \textit{vector2vector (v2v)}—to supplement standard loss functions and enhance the effectiveness of KD in molecular GNNs. These strategies help bridge the gap between teacher and student models, improving knowledge transfer in complex molecular systems.

The standard loss function $\mathcal{L}_0$ for molecular GNNs, as outlined in Eq. \ref{eq:loss_energy_force}, accounts for both energy and force predictions:

\begin{equation}
\mathcal{L}_0=\alpha_{\mathrm{E}} \mathcal{L}_{\mathrm{E}}(\hat{E}, E)+\alpha_{\mathcal{F}} \mathcal{L}_{\mathrm{F}}(\hat{\boldsymbol{F}}, \boldsymbol{F})
\label{eq:loss_energy_force}
\end{equation}

where $E$ and $F$ represent the ground-truth energy and forces, while $\hat{E}$ and $\hat{F}$ denote their predicted counterparts. The terms $\mathcal{L}_E$ and $\mathcal{L}_F$ denote the energy and force loss functions, respectively, weighted by $\alpha_{\mathrm{E}},\,\alpha_{\mathrm{F}}\in\mathbb{R}$.

For knowledge distillation \cite{hinton2015distilling}, the loss function in Eq. \ref{eq:loss_energy_force} is augmented with an auxiliary distillation loss $\mathcal{L}_{\mathrm{KD}}$, resulting in the following formulation:
\begin{equation*}
\mathcal{L}=\mathcal{L}_0+\lambda \mathcal{L}_{\mathrm{KD}}.
\label{eq:aux_loss}
\end{equation*}

In this work, we aim to develop a more efficient variant of the foundational model JMP-L without compromising its performance. Since JMP is built on GemNet-OC~\cite{gasteiger2022gemnet}, we focus on the \textit{node2node} and \textit{edge2edge} losses as key components of the distillation process. Unlike previous studies that apply distillation only at the final task level, our primary objective is to enhance the efficiency of the foundational model itself while systematically assessing its impact on downstream tasks. Specifically, we examine how distillation influences the model’s generalization capabilities, providing deeper insights into its performance across diverse molecular property prediction tasks.

\subsubsection{Pruning Coupled with Distillation} 
Aggressive structured pruning can substantially degrade model performance. For instance, brute-force structural pruning methods, such as L2-based filter-wise pruning, have led to a 50-fold performance drop in LLMPruner \cite{ma2023llmpruner}. However, in large models, aggressive pruning combined with fine-tuning can considerably reduce the number of layers—sometimes by half—while incurring only minimal performance loss \cite{gromov2024unreasonableineffectivenessdeeperlayers}. Techniques such as parameter-efficient fine-tuning, quantization, and low-rank adapters further help preserve model accuracy post-pruning \cite{gromov2024unreasonableineffectivenessdeeperlayers}. Recent KD approaches, such as those proposed by Ekström \textit{et al.} \cite{ekstrom2024accelerating}, require training the student model from scratch, demanding substantial computational resources. To the best of our knowledge, no prior work has explored the combined use of pruning and distillation for DFT and MD molecular property prediction. Our approach applies distillation to a pre-trained, block-reduced network, offering the potential for improved accuracy while greatly reducing both training and inference time.

\section{Block Reduction for Efficient Foundation Models}

\begin{figure*}[t]
    \centering
    \includegraphics[width=\textwidth]{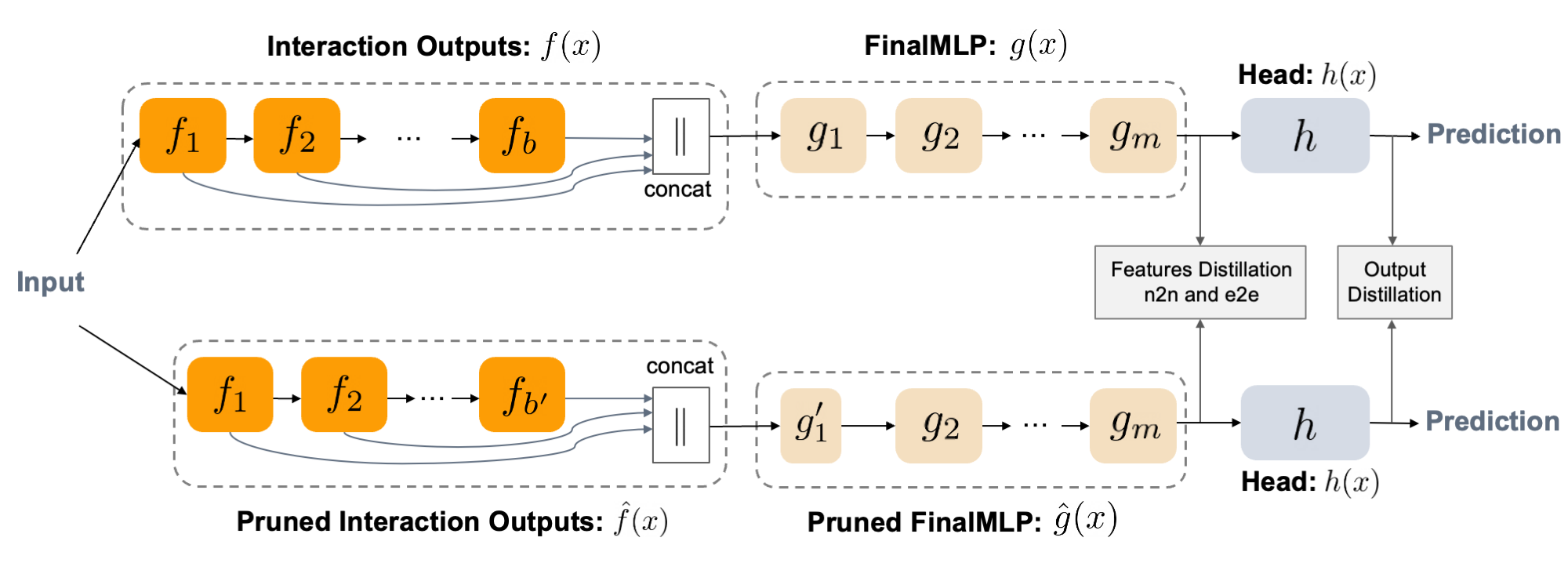}
    \vspace{-0.8cm}
    \caption{\textbf{Block Reduction for Efficient Foundation Models.} The top model represents the foundation model JMP-L, where interaction blocks extract features, which are concatenated and processed by FinalMLP before making predictions. The bottom model is its pruned version, constructed by removing low-importance blocks and adjusting FinalMLP. To mitigate performance degradation, we apply both feature distillation (node-to-node and edge-to-edge) and output distillation to transfer knowledge from the original model.}
    \vspace{-0.2cm}
    \label{fig:pipeline}
\end{figure*}

\subsection{Preliminaries}
Our work is based on the GemNet-OC architecture~\cite{gasteiger2022gemnet}, although our analysis can be applied to similar architectures. 

In particular, we define $f_{\theta}:\mathbb{R}^{4\times n}\rightarrow \mathbb{R}^{n\times d}$ as a function that maps a molecular graph, represented by the 3D positions and atomic numbers of $n$ atoms, to a feature space. The feature extraction process is formulated as follows:

\begin{equation}
\begin{aligned}
f(x) = \text{concat}(&f_1(x), f_2 \circ f_1 (x), \dots, \\
& f_b \circ f_{b-1} \circ \dots \circ f_2 \circ f_1 (x))
\end{aligned}
\label{eq:blocks}
\end{equation}

where the model extracts features through $b$ sequential blocks. Each block $f_i: \mathbb R^{n\times d}\rightarrow \mathbb R^{n\times d}$ (for $i>1$) refines the representations, while the initial block $f_1: \mathbb R^{n} \rightarrow \mathbb R^{n\times d}$ is an embedding layer that performs the initial transformation. The resulting feature space is of dimension $\mathbb{R}^{n\times (d\times b)}$, obtained by concatenating outputs from all $b$ blocks. This extracted feature representation is then processed through a sequence of multilayer perceptron (MLP) layers, known as FinalMLP in GemNet-OC.
\begin{equation*}
g(x) = g_m \circ g_{m-1} \circ \dots \circ g_2 \circ g_1(x)
\end{equation*}
where $g_1: \mathbb{R}^{n \times (d \times b)} \to \mathbb{R}^{n \times d}$ transforms the concatenated features into $d$, with $g(x): \mathbb{R}^{n \times (d \times b)} \to \mathbb{R}^{n \times d}$ providing the final transformation. Finally, the output of $g(x)$ is passed to a prediction head $h(x)$, which predicts the three-dimensional force vector for each atom and the molecule’s energy.
The full model is the composition between the feature extractor and the MLP layers given by the following:
\begin{equation}
F(x) = h\circ g\circ f(x).
\label{eq:final_output}
\end{equation}
For the GemNet-OC architecture used by JMP-L, $b=7$ (one embedding layer and six interaction blocks) and $m=5$, resulting in $160.1$M parameters in total. JMP-L is shown on top in Figure \ref{fig:pipeline}.

\subsection{Interaction Block Importance}
\label{sec:interactionblock}
Interaction blocks are a fundamental component of machine learning potential models (e.g., SchNet~\cite{schutt2017schnet}, GemNet-OC~\cite{gasteiger2021gemnet}), enabling richer representations and capturing long-range atomic interactions. These models stack multiple interaction blocks in a sequential manner to build higher-body representations and model complex atomic relationships effectively. However, quantifying each interaction block's contribution to the final prediction is not straightforward, as interactions are highly interdependent and difficult to isolate.

To address this, we propose an approach to measure interaction block importance within the GemNet-OC backbone used by JMP-L. We employ GradCAM~\cite{gradcam} to assess each block’s impact on the final output. First, we extract and concatenate the output features from all $b$ blocks, as formulated in Eq.\ref{eq:blocks}. Using these features ($f$) and Eq.~\ref{eq:final_output}, we compute the model’s output and its corresponding loss, $\mathcal{L}_0$. We then compute the gradient $\nabla_{\text{CAM}}$ of $f$ with respect to $\mathcal{L}_0$ and determine each block’s relevance $r$ using:
\begin{equation*} r = \text{ReLU}\left(f \circ \nabla_{\text{CAM}} \right). 
\end{equation*}

Following the GradCAM methodology, we apply a ReLU activation to emphasize features that positively contribute to the model’s prediction. To quantify the contribution of each interaction block, we decompose the relevance map $r$, which represents the overall importance of features, into $b$ partitions. Each partition $r_i$ corresponds to a feature dimension $d$ of a specific interaction block. Finally, we compute the importance score for each block by averaging $r_i$ across both the feature and batch dimensions, providing a measure of its overall contribution to the final prediction.

After using GradCAM, we normalize the contributions by dividing each individual contribution by the sum of all contributions. This ensures that the importance scores are comparable across different models and configurations.

\subsection{Block Reduction Strategies}
Although the pre-trained GemNet-OC (JMP-L) achieves strong performance when fine-tuned across various tasks and datasets, fine-tuning it remains computationally expensive. This inefficiency arises from its large architecture and the high computational cost of each forward pass. For instance, fine-tuning JMP-L on rMD17—containing only 1,000 graphs—still requires 160.1M parameters. Given the small scale of rMD17, this parameter count appears disproportionately large, limiting the practicality of leveraging such a powerful pre-trained model efficiently.

In this work, we aim to improve the efficiency of foundation pre-trained models through block reduction. Specifically, we explore different strategies to construct a reduced model $\hat F (x) = h \circ \hat g \circ \hat f(x)$, where:

\begin{align*}    
\hat f(x) = \text{concat}(&f_1(x), f_2 \circ f_1 (x), \dots,\\ &f_{b'} \circ f_{b'-1} \circ \dots \circ f_2 \circ f_1 (x))
\end{align*}
and
\begin{equation*}
\hat g(x) = g_{m} \circ g_{m-1} \circ \dots \circ g_2 \circ g'_1(x)
\end{equation*}
with $g'_1: \mathbb R^{n\times (d\times b')}\rightarrow \mathbb R^{n\times d}$ and $b' < b$. 

This formulation reduces the original architecture by removing the last 
$b - b'$ interaction blocks and adjusting the dimensionality of the first MLP block ($g'_1$) accordingly. However, removing interaction blocks disrupts the alignment between the feature extractor $\hat f$ and the FinalMLP $\hat g$, as it alters the structure of the extracted features. To address this misalignment, we explore three main strategies to restore compatibility between the reduced feature extractor and the FinalMLP.

\paragraph{Random MLP.}
Random MLP is the simplest baseline, where we resize the first MLP layer $g'_1$ and randomly initialize its weights. This approach assumes that the features extracted by the remaining interaction blocks are still useful and that the weights of $g'_1$ can be learned effectively during fine-tuning. We refer to this strategy as RandomMLP in the experimental section.

\paragraph{Sliced MLP.}
In the Sliced MLP strategy, we retain the parameters of the first MLP layer $g'_1$  from the original model, truncating it to match the reduced dimensionality of the features. This assumes that the preserved parameters provide a good initialization for fine-tuning, maintaining continuity between the pre-trained and pruned model. Unless otherwise stated, all of our block reduction experiments follow this strategy. 

\paragraph{Knowledge Distillation.}
In the knowledge distillation approach, we introduce a learning paradigm tailored for block reduction. Specifically, we follow \cite{ekstrom2024accelerating} by distilling the force predictions of the pre-trained model $F$ into its block-reduced counterpart $\hat F$. This is achieved by optimizing the following objective:
\begin{equation*}
\min_\theta \mathbb E_{x\sim \mathcal D} \|\hat F_\theta(x) - F_{\theta_0}(x)\|_1
\end{equation*}
where $\mathcal D$ represents the data distribution used during the pre-training phase of $F$.

To further align the representations between the original and pruned models, we incorporate node-to-node (n2n) and edge-to-edge (e2e) distillation. This extends the objective to:

\begin{align*}
\min_{\theta} \mathbb{E}_{x \sim \mathcal{D}} \Big[ & \| \hat{F}(x; \theta) - F(x; \theta_0) \|_1 + \\
& \sum_{i=1}^{n'} \| \hat{g}_i(\hat{f}(x); \theta) - g_i(f(x); \theta_0) \|_1 \Big].
\end{align*}

The first term represents output distillation, while the second ensures feature-level consistency between the pruned and original models. Our full pipeline is illustrated in Figure \ref{fig:pipeline}.

It is important to note that all prior block reduction approaches operate on the pre-trained foundational model. Thus, block reduction produces a generalist model, which must still undergo fine-tuning on task-specific datasets to become a specialized model for a given downstream application.

\section{Experiments}
\label{sec:experiments}

\subsection{Deeper layers contribute less}
\begin{figure}[!ht]   
    \centering
    \includegraphics[width=\columnwidth]{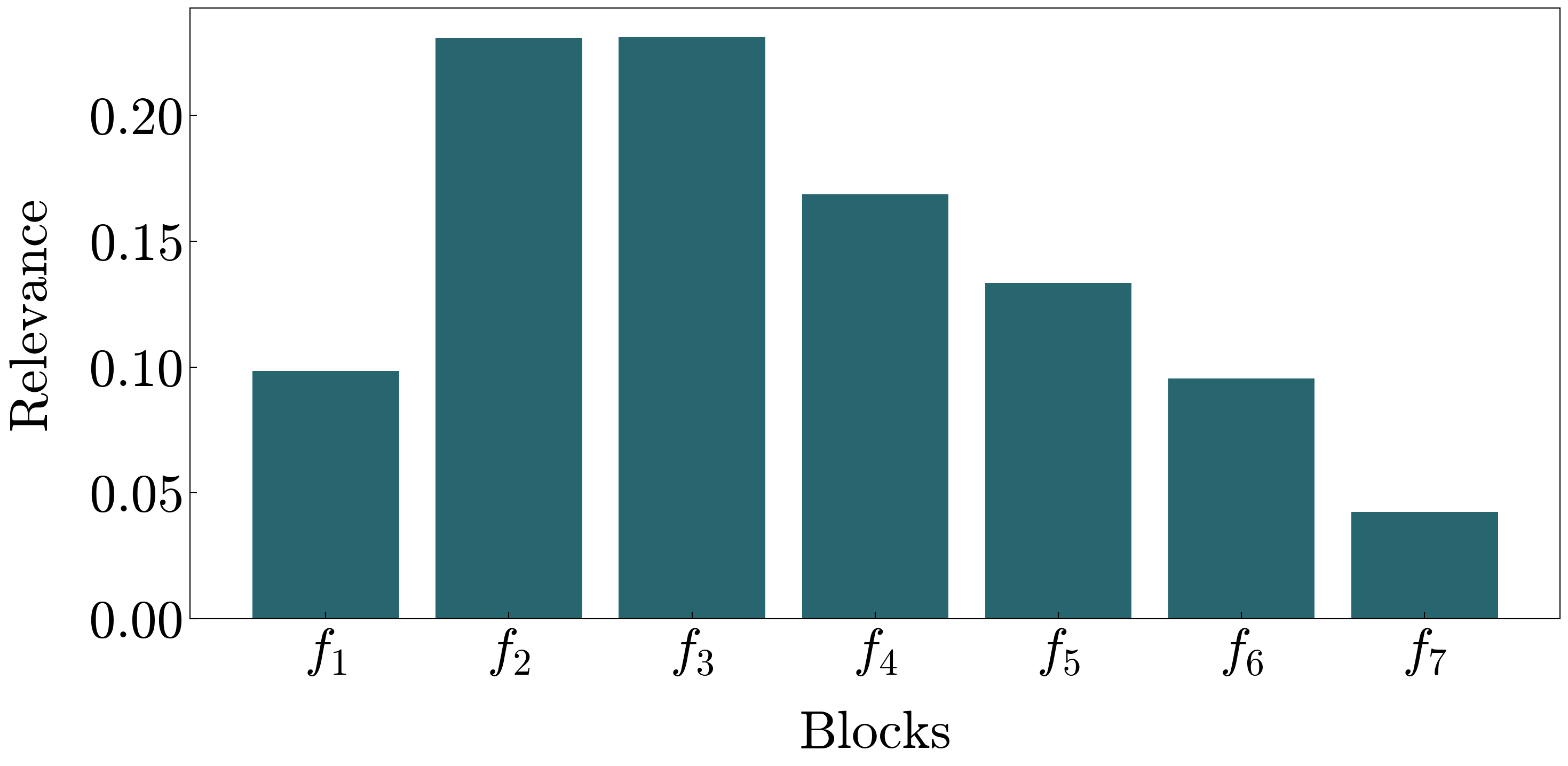}
    \vspace{-0.6cm}
    \caption{\textbf{Block Relevance Analysis.}  This table illustrates the contribution of each output block in JMP-L to the final prediction. The first block, $f_1$, represents the embedding output, while $f_2$ to $f_7$ correspond to the six interaction blocks of GemNet-OC. The results show diminishing returns in deeper interaction blocks, suggesting they are strong candidates for pruning to improve efficiency without considerable performance loss.}
      \label{fig:block_imp}
\end{figure}

To develop a more efficient pre-trained version of JMP-L for diverse downstream tasks, we analyze the contribution of each interaction block to the final output prediction on the pre-training distribution, as outlined in Sec.\ref{sec:interactionblock}. Figure~\ref{fig:block_imp} reveals a gradual decline in relevance in deeper layers, with the sixth and seventh blocks exhibiting the lowest contribution. This gradual decline suggests that blocks with low contribution can be removed with minimal impact on performance. Although the embedding block ($f_1$) also shows low relevance, it directly feeds into the first interaction block ($f_2$), which initializes message passing and is crucial for propagating information between atoms and bonds. Removing the embedding block could introduce major structural disruptions, potentially degrading performance, and is therefore not considered for block reduction.

\begin{table*}[t]
\caption{\textbf{Force MAE Evaluation During Pre-training.} Impact evaluation of our block reduction (BR) approach on pre-training performance, comparing results with and without knowledge distillation (KD). The reported values represent the force MAE (meV/Å) across the OC20, OC22, ANI-1x, and Transition-1x datasets used for pre-training. KD reduces the performance drop relative to the teacher model. Note: The block count excludes the embedding output.}
    \centering
    \renewcommand{\arraystretch}{1.3} 
    \begin{tabular}{@{}lccccc@{}}
        \toprule
         & \# of Blocks & OC20 & OC22  & ANI-1x & Transition-1x \\ \midrule
        BR & 2 & 106.6 (\textcolor{negative}{-89.6}) & 111.9 (\textcolor{negative}{-89.0}) & 711.8 (\textcolor{negative}{-689.3}) & 188.3 (\textcolor{negative}{-175.5}) \\
        \hspace{1em}+ KD & 2 & 52.1 (\textcolor{positive}{-35.1}) & 56.3 (\textcolor{positive}{-33.4}) & 144.4 (\textcolor{positive}{-121.9}) & 53.3 (\textcolor{positive}{-40.6}) \\
        \midrule
        BR & 3 & 94.1 (\textcolor{negative}{-77.1}) & 99.3 (\textcolor{negative}{-76.3}) & 615.1 (\textcolor{negative}{-592.6}) & 149.6 (\textcolor{negative}{-136.9}) \\
        \hspace{1em}+ KD & 3 & 42.1 (\textcolor{positive}{-25.2}) & 45.6 (\textcolor{positive}{-22.7}) & 97.6 (\textcolor{positive}{-75.1}) & 39.1 (\textcolor{positive}{-26.4}) \\
        \midrule
        BR & 4 & 65.3 (\textcolor{negative}{-48.4}) & 68.9 (\textcolor{negative}{-46.0}) & 443.4 (\textcolor{negative}{-421.0}) & 99.8 (\textcolor{negative}{-87.1}) \\
        \hspace{1em}+ KD & 4 & 26.7 (\textcolor{positive}{-9.8}) & 34.5 (\textcolor{positive}{-11.6}) & 58.1 (\textcolor{positive}{-35.7}) & 24.2 (\textcolor{positive}{-11.4}) \\
        \midrule
        BR & 5 & 39.1 (\textcolor{negative}{-22.1}) & 45.7 (\textcolor{negative}{-22.7}) & 220.0 (\textcolor{negative}{-197.6}) & 48.5 (\textcolor{negative}{-35.8}) \\
        \hspace{1em}+ KD & 5 & 19.8 (\textcolor{positive}{-2.8}) & 25.8 (\textcolor{positive}{-2.9}) & 30.1 (\textcolor{positive}{-7.7}) & 16.0 (\textcolor{positive}{-3.3}) \\
        \midrule
        JMP-L (Teacher) & 6 & 17.0 & 22.9 & 22.5 & 12.7 \\
        \bottomrule
    \end{tabular}
    \label{tab:pretraining_metrics}
\end{table*}

\subsection{Block Reduction and Distillation in Pre-training}
\label{section_4_2}
We investigate the effectiveness of block reduction (BR) and knowledge distillation (KD) during the pre-training phase. The importance analysis of the blocks shows that the deeper interaction blocks in JMP-L contribute less to the final prediction compared to the earlier ones. This raises a key question: how much do these later blocks impact prediction performance, and if their removal leads to degradation, to what extent can knowledge distillation recover the lost performance? To address these questions, we explore a combined approach of block reduction and knowledge distillation, assessing whether distilling knowledge from the full model into a block-reduced version can maintain performance while reducing computational costs.

\noindent\textbf{Settings:}
We use the same pre-training datasets as in JMP~\cite{shoghi2023molecules}, including OC20~\cite{chanussot2021open}, OC22~\cite{tran2023open}, ANI-1x~\cite{smith2020ani}, and Transition-1x~\cite{schreiner2022transition1x}, with a total of 120M training samples. Following our block reduction (BR) strategy, we sequentially remove interaction blocks starting from the last one, as indicated by the importance analysis in the previous section. This allows us to construct progressively smaller versions of JMP-L, retaining 5, 4, 3, and 2 interaction blocks. To mitigate potential performance degradation, we apply a brief knowledge distillation (KD) phase to each pruned model using less than 1.5\% of the pre-training datasets. We find that running KD for under 2 GPU-days on an A100 is sufficient for convergence. We report performance using the mean absolute error (MAE).

\noindent\textbf{Observations:} 
Table~\ref{tab:pretraining_metrics} presents the results of block reduction (BR) and knowledge distillation (KD) during the pre-training stage. Removing interaction blocks leads to a performance drop proportional to the number of blocks removed. Notably, OC20 and OC22 show smaller performance drops compared to other datasets, likely due to their higher force loss weight during pre-training~\cite{shoghi2023molecules}, making the model less sensitive to deeper block removal. Applying both BR and KD greatly reduces the performance gap with the teacher model. For instance, with 5 blocks, the performance difference narrows to just -2.8 and -2.9 meV/Å for OC20 and OC22, respectively. These results demonstrate the effectiveness of combining BR and KD in maintaining predictive accuracy during pre-training. Next, we analyze how well these pruned models perform on downstream tasks.

\subsection{Main Results}
\label{sec:main_results}
We evaluate pruned versions of JMP-L across various downstream tasks and knowledge transfer strategies to determine the most efficient fine-tuning approach for optimal performance. A key question is whether distilling during pre-training is more effective than applying block reduction alone. To explore this, we consider different baselines during fine-tuning.

\noindent\textbf{Baselines:} 
We evaluate the following fine-tuning strategies:
\begin{itemize}[leftmargin=1em, topsep=0pt, itemsep=0pt, parsep=0pt]
    \item BR (Block Reduction): Remove interaction blocks and slice their corresponding weights in the first layer of FinalMLP.
    \item BR/RandomMLP: A simpler variant of BR, where instead of pruning the first layer of FinalMLP, we randomly initialize a smaller version to match the reduced number of interaction blocks.
    \item BR+KD (Block Reduction + Knowledge Distillation): Load the pruned and distilled version of JMP-L, where knowledge distillation has been applied during pre-training.
\end{itemize}

We further report the original performance of JMP-L from~\cite{shoghi2023molecules} alongside our reproduced version. 
Our fine-tuning process is constrained by a fixed computational budget, compared to the vanilla JMP-L:

\begin{figure*}[!ht]   
    \centering
    \includegraphics[width=\textwidth]{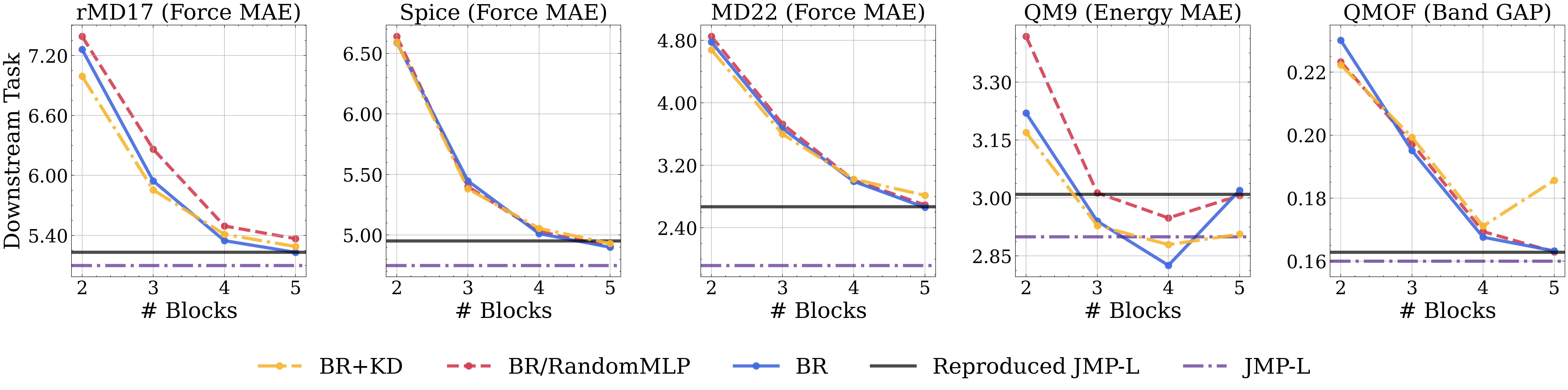}
    \vspace{-0.6cm}
    \caption{\textbf{Evaluation on downstream tasks.} 
    Evaluation of the performance across various downstream tasks using different block reduction strategies: block reduction (BR), block reduction with a randomly initialized MLP (BR/RandomMLP), and block reduction combined with knowledge distillation (BR+KD). Performance is measured in MAE: meV/Å for force targets, meV for the QM9 energy target, and eV for the QMOF band gap target. The original JMP-L model utilizes 6 blocks. 
    } 
      \label{fig:main_results}
\end{figure*}

\noindent\textbf{Settings:}
We evaluate the baselines on a representative set of targets from the datasets used in~\cite{shoghi2023molecules}, specifically: Aspirin (rMD17~\cite{chmiela2017machine}), $U_0$ (QM9~\cite{ramakrishnan2014quantum}), Solvated Amino Acids (SPICE~\cite{eastman2023spice}), Ac-Ala3-NHMe (MD22~\cite{chmiela2023accurate}), and Band Gap (QMOF~\cite{rosen2021machine}). 
For a fair comparison, since the baselines differ in computational demands, we fine-tune each model for 1 GPU-day on a V100, except for QM9, which requires 2 GPU-days to approach convergence. We then evaluate the models on the test set of the corresponding dataset and target. Our results are shown in Figure~\ref{fig:main_results}.

\noindent\textbf{Block Reduction is a Strong Baseline:}
Figure~\ref{fig:main_results} highlights the surprising effectiveness of BR (shown in blue) across datasets. In particular, the 5-block model matches the performance of the 6-block reproduced JMP-L baseline. Even the 4-block model remains competitive across most tasks and outperforms the original JMP-L on the QM9 target. These results suggest that for tasks with longer convergence times, such as QM9, a compressed model not only reduces computational costs but may also converge faster and even surpass the full model’s performance. However, when further reducing to 3 or 2 blocks, we observe a more pronounced drop in performance, indicating that the model may be underfitting the task.

\begin{figure*}[!ht]   
    \centering
    \includegraphics[width=\textwidth]{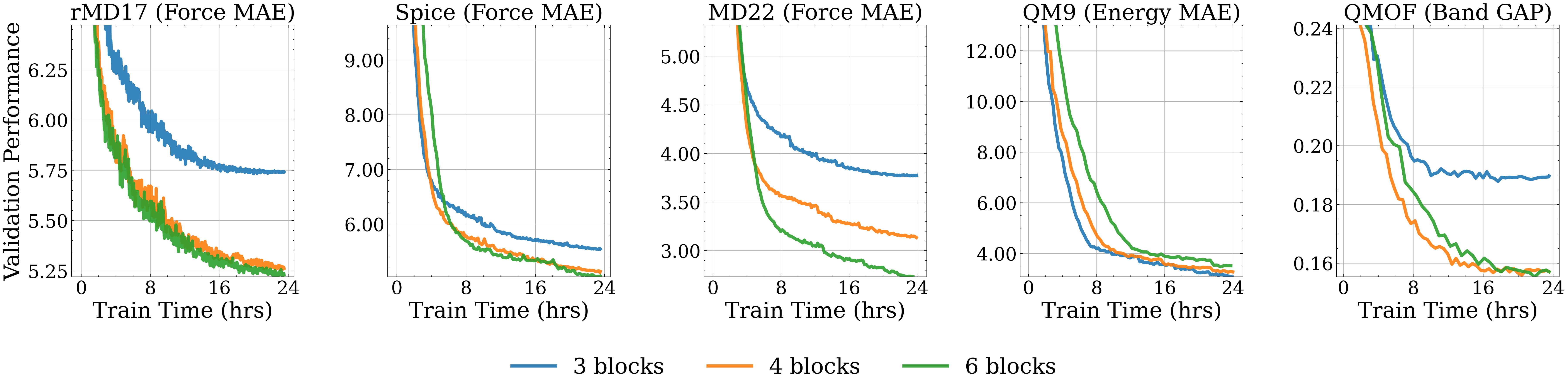}
    \vspace{-0.6cm}
    \caption{
        \textbf{Training Efficiency Analysis.} We compare the convergence speed of JMP-L models with 3, 4, and 6 blocks over a fixed 1 GPU-day training budget. The 4-block model achieves faster convergence than the 6-block model on QM9 after 8 hours but lags behind on MD22, while performing comparably on other datasets. Performance is measured in MAE: meV/Å for force targets, meV for the QM9 energy target, and eV for the QMOF band gap target.
    }
    \label{fig:training_speed}
\end{figure*}

\noindent\textbf{Pre-training Distillation Works in Certain Scenarios:} 
While KD improved performance on the pre-training datasets (as discussed in Section~\ref{section_4_2}), its effectiveness on downstream tasks varies, as shown by BR+KD (yellow in Figure~\ref{fig:main_results}). For example, KD improves performance on rMD17 when using 2 or 3 blocks, but it hurts the performance with 4 and 5 blocks. This could be due to the fact that the last two blocks of JMP-L contribute less to downstream tasks (as indicated by BR results), meaning that distilling from these less relevant blocks during pre-training may introduce noise and distort useful features. Interestingly, KD improves performance for the 5-block model in QM9, suggesting a potential edge case where distillation benefits from specific task characteristics.

\noindent\textbf{JMP-L's FinalMLP Layer May Indicate Distribution Shifts:}
The BR/RandomMLP baseline (red in Figure~\ref{fig:main_results}) exhibits inconsistent behavior across different tasks. In QMOF, SPICE, and MD22, randomly initializing the first layer of FinalMLP had little impact on performance, suggesting a distribution shift between pre-training and downstream tasks. In contrast, BR/RandomMLP shows a noticeable performance drop in rMD17 and QM9, indicating that the learned FinalMLP features are more relevant to these tasks. This observation also aligns with the improved performance of BR+KD in rMD17 and QM9, where KD effectively preserves useful representations. These findings suggest that FinalMLP features could serve as indicators of distribution differences across tasks, highlighting variations in task similarity to the pre-training distribution.

\subsection{Evaluating Training and Inference Efficiency}

\noindent\textbf{Training Efficiency:}
We complement our analysis with both training and inference speed-ups gains. For training efficiency, Figure~\ref{fig:training_speed} illustrates the convergence speed over a fixed budget of 1 GPU-day for models with 3, 4, and 6 blocks across five datasets: rMD17, SPICE, MD22, QM9, and QMOF. The performance is measured in force MAE (meV/Å) for rMD17, SPICE, and MD22, energy MAE (meV) for QM9, and band gap error (eV) for QMOF.

In general, reducing the number of interaction blocks accelerates training. The 4-block model has similar convergence behavior to the full 6-block model across most datasets. For example, on QM9, the 4-block model outperforms the 6-block model after 8 hours of training, achieving lower energy MAE. However, on MD22, the 4-block model lags behind, with the 6-block model maintaining a noticeable performance advantage. In contrast, the 3-block model shows the fastest convergence across datasets but exhibits a worse final validation loss, indicating a trade-off between efficiency and accuracy.

These findings emphasize the importance of selecting an optimal block count to balance convergence speed and final model accuracy. The 4-block model strikes a balance, providing a moderate training speedup over the 6-block model in some datasets while maintaining competitive performance across all datasets.

\begin{table}[h]
    \centering
\caption{
    \textbf{Inference Efficiency Analysis.} We evaluate the impact of block reduction on JMP-L's efficiency using a subset of the QMOF validation set. Reducing interaction blocks lowers computational cost and improves inference speed. The 4-block model provides the best trade-off, achieving a 1.3x speedup with a 32\% reduction in parameters.
}
    \renewcommand{\arraystretch}{1.2}
    \resizebox{\linewidth}{!}{
    \begin{tabular}{lccc} 
        \toprule
        \makecell{\bf \centering Blocks} & \makecell{ \textbf{Throughput} \\ \textbf{(samples/s)}} & \makecell{ \textbf{GFlops} \\ \textbf{(Billion)}} & \makecell{ \textbf{Parameters} \\ \textbf{(M)}}
        \\\midrule
        6-blocks & 19.1 & 1.74 & 160.9
        \\\midrule
        5-blocks & 21.8 (\textcolor{positive}{+2.7}) & 1.45 (\textcolor{negative}{-0.29}) & 134.5 (\textcolor{negative}{-16\%})
        \\
        4-blocks & 25.6 (\textcolor{positive}{+6.5}) & 1.16 (\textcolor{negative}{-0.58}) & 108.2 (\textcolor{negative}{-32\%})
        \\
        3-blocks & 30.8 (\textcolor{positive}{+11.7})& 0.87 (\textcolor{negative}{-0.87}) & 81.9 (\textcolor{negative}{-49\%})
        \\
        2-blocks & 38.0 (\textcolor{positive}{+18.9})& 0.59 (\textcolor{negative}{-1.15}) & 55.5 (\textcolor{negative}{-65\%})
        \\
        \bottomrule
    \end{tabular}
    }
    \label{tab:inference_table}
\end{table}

\noindent\textbf{Inference Efficiency:}
To assess inference efficiency, we use the QMOF dataset, which has a large average graph size, making it a representative benchmark for evaluating computational cost. Using a V100 GPU, we evaluate inference throughput on a subset of the QMOF validation set and report results in Table~\ref{tab:inference_table}.

Reducing the number of interaction blocks improves inference throughput by lowering computational overhead. The 6-block model processes 19.1 samples per second, while reducing to 5 blocks increases throughput to 21.8 samples per second, reflecting a moderate efficiency gain. Further reductions yield more substantial improvements: the 4-block model achieves 25.6 samples per second, providing a 1.3× speedup over the 6-block model while maintaining strong predictive performance. The 3-block and 2-block models offer even higher throughput, but at the cost of accuracy loss as seen in Figure~\ref{fig:main_results}.

Our block reduction strategy greatly lowers computational cost, as reflected in GFlops and parameter count. The 4-block model reduces GFlops from 1.74B to 1.16B and parameters from 160.9M to 108.2M (-32\%), demonstrating its efficiency gains. This targeted reduction makes the block-reduced models particularly well-suited for deployment in resource-constrained environments and high-throughput screening applications, where maintaining predictive performance while optimizing computational efficiency is essential.

\section{Generalization to Other Architectures}
Our analysis in Section~\ref{sec:experiments} is based on the GemNet-OC~\cite{gasteiger2022gemnet} architecture within the JMP~\cite{shoghi2023molecules} pre-training paradigm. To the best of our knowledge, JMP is the only large-scale, multi-domain pre-training approach for atomic property prediction. While several architectures provide pre-trained checkpoints on the OC20 catalysis dataset~\cite{chanussot2021open}, it is unclear whether these models generalize as well as JMP across diverse downstream tasks.

To evaluate the generalizability of our block reduction methodology beyond GemNet-OC, we extend our study to EquiformerV2~\cite{liao2023equiformerv2}, a state-of-the-art transformer-based equivariant model for atomic property prediction. Using the same JMP pre-training methodology, we pretrain EquiformerV2 with 31M parameters on the same four upstream tasks. Due to computational constraints, we limit pre-training to 12 A100 GPU days. We then apply our block reduction and knowledge distillation strategies and evaluate the model across multiple downstream tasks. While block reduction and knowledge distillation considerably reduced training time with minimal performance loss for JMP-L (160M parameters), their impact on EquiformerV2 (31M parameters) was more limited. This suggests that our approach is most effective for larger models, where over-parameterization leaves room for optimization. We include the evaluation results and discussion in the Appendix.

\section{Conclusion}
In this work, we explored strategies to enhance the efficiency of foundation models for atomic property prediction. By analyzing the role of individual layers in JMP-L, we found that deeper interaction blocks contribute less to predictive accuracy, making them suitable candidates for pruning. Our results show that reducing JMP-L’s parameter count by 32\% improves inference throughput by 1.3× while maintaining comparable performance. Additionally, we demonstrated that knowledge distillation can help mitigate performance degradation in certain tasks.
Our methodology reduces inference time and computational cost, making the optimized models practical for scientists conducting research in resource-constrained settings and high-throughput screening, where both accuracy and efficiency are crucial.
We hope this study encourages further research into efficient training and inference for molecular property prediction, paving the way for lighter models in molecular and materials discovery.

\vspace{2pt}\noindent\textbf{Acknowledgments.}
This work is supported by the KAUST Center of Excellence for Generative AI under award number 5940. The computational resources are provided by IBEX, which is managed by the Supercomputing Core Laboratory at KAUST. Yasir is supported by Saudi Aramco.

{
    \small
    \bibliographystyle{ieeenat_fullname}
    \bibliography{main}
}

\clearpage
\appendix
\setcounter{figure}{0}
\setcounter{table}{0}
\section*{Appendix}

\section{Beyond GemNet-OC}
In this section, we extend our block reduction and knowledge distillation methodology to EquiformerV2~\cite{liao2023equiformerv2}, a state-of-the-art transformer-based equivariant model for atomic property prediction. Following the same pipeline as in the main paper, we first apply GradCAM as described in Section~\ref{sec:interactionblock} to assess interaction block importance. We then pre-train EquiformerV2 following the same methodology used for JMP before evaluating its performance on the downstream tasks.

\subsection{Block Relevance}
Using the JMP~\cite{shoghi2023molecules} pre-training datasets (OC20, OC22, ANI-1x, and Transition-1x), we perform forward passes on 1,000 samples from each dataset to extract features from the EquiformerV2 backbone (31M parameters). This backbone consists of 8 transformer blocks. We save the output features of each block and apply GradCAM (as described in Section~\ref{sec:interactionblock}) to assess the relevance of each block to the final output prediction.

\begin{figure}[!ht]   
    \centering
    \includegraphics[width=\columnwidth]{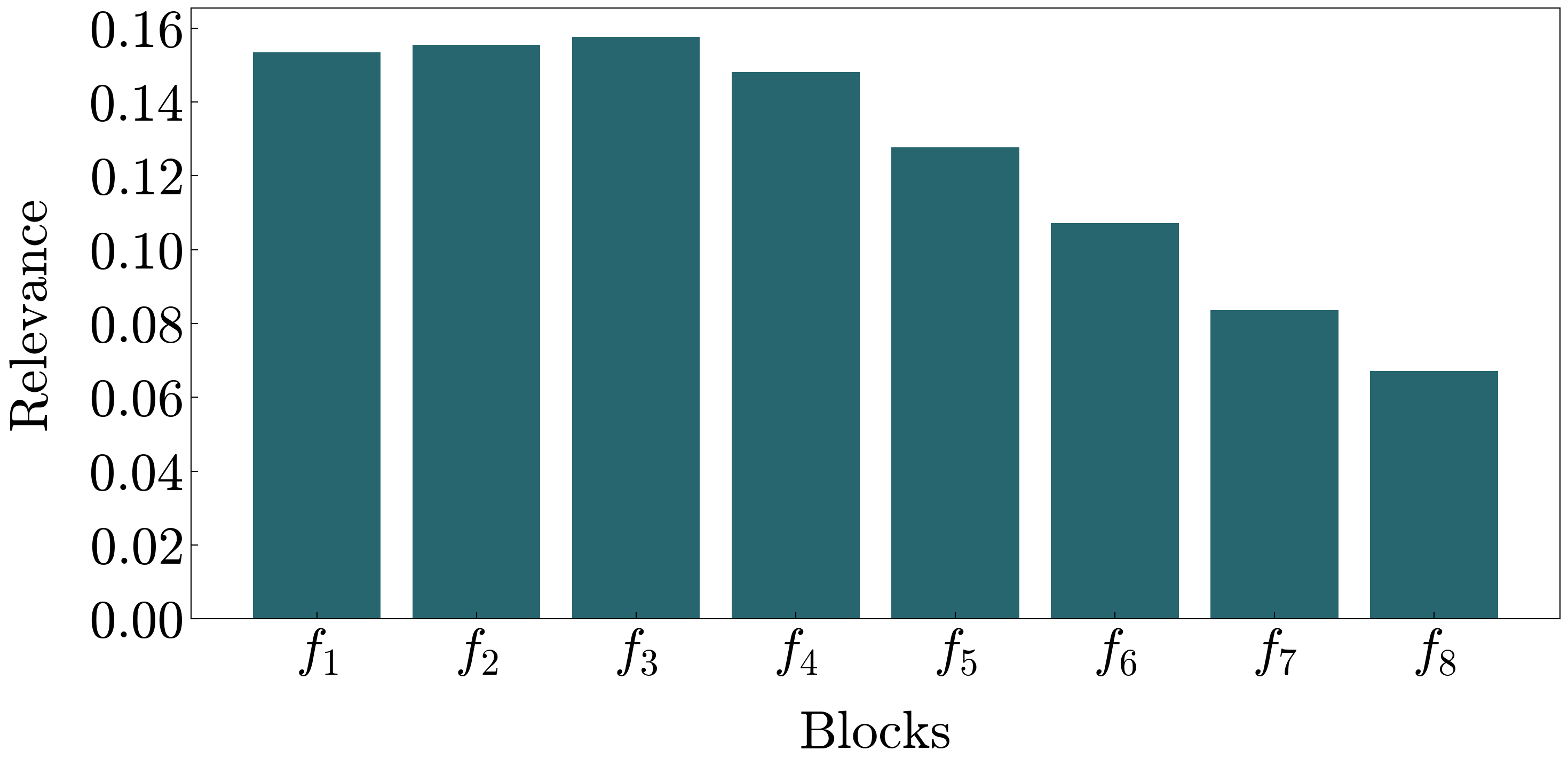}
    \caption{\textbf{Block Relevance Analysis for EquiformerV2.} This figure illustrates the contribution of each transformer block in EquiformerV2 to the final prediction. The model consists of eight transformer blocks, $f_1$ to $f_8$, with relevance decreasing in the deeper layers. The results suggest that later blocks are strong candidates for pruning to improve efficiency with minimal performance loss.}
    \label{fig:eqv2_block_imp}
\end{figure}

Appendix Figure~\ref{fig:eqv2_block_imp} presents the relevance scores of EquiformerV2’s transformer blocks. Similar to GemNet-OC, we observe a diminishing return in later blocks, where deeper layers contribute less to the final prediction. This trend suggests that the later transformer blocks are strong candidates for pruning, reinforcing the effectiveness of our block reduction strategy for improving model efficiency.

\subsection{Block Reduction and Distillation in Pre-training for EquiformerV2}
\textbf{Pre-training setting:} While GemNet-OC has publicly available pre-trained weights on all four JMP datasets, EquiformerV2 only provides pre-trained weights for OC20 or OC22 individually. To ensure a fair and consistent comparison, we initialize the EquiformerV2 backbone with the FAIRChem pre-trained weights (EquiformerV2-31M-S2EF-OC20-All+MD). We then pre-train the model following the JMP pre-training paradigm, using a uniform balancing strategy with 2M samples from each upstream dataset, totaling 8M samples. The pre-training is conducted over 2 epochs using 4 A100 GPUs, with a total compute time of 12 A100 GPU-days.

After pre-training, we apply block reduction (BR) and knowledge distillation (KD) following the approach outlined in Section~\ref{section_4_2} of the main paper. For block reduction, we sequentially remove later blocks, constructing smaller models with 7, 6, 5, and 4 transformer blocks. Since removing blocks may lead to performance degradation, we conduct a brief KD phase to recover performance for each pruned model. To manage computational cost, we limit the KD phase to 1.5\% of the pre-training size. This process requires approximately 63 A100 GPU-hours.

\noindent\textbf{Observations:} Appendix Table~\ref{tab:pretraining_metrics_EqV2} presents the force MAE performance across various pre-training tasks for EquiformerV2. Consistent with GemNet-OC results in the main paper, we observe that removing later blocks leads to a proportional performance drop, confirming that deeper layers contribute less to the final prediction. Applying knowledge distillation after block reduction considerably improves performance, mitigating much of the accuracy loss. For example, in the 4-block model, KD reduces the performance gap with the teacher model from -50.9 meV/Å to -5.9 meV/Å for OC20 and from -40.4 meV/Å to -10.3 meV/Å for OC22. Similar gains are observed for ANI-1x and Transition-1x, highlighting the effectiveness of KD in compensating for the loss of capacity from pruning.

These results reinforce the effectiveness of our block reduction and KD approach across different model architectures, demonstrating that smaller, more efficient models can retain strong predictive accuracy with minimal fine-tuning overhead. Next, we evaluate the performance of these pruned models on downstream tasks.

\begin{table*}[h]
    \caption{\textbf{Force MAE Evaluation During Pre-training of EquiformerV2.} We analyze the impact of block reduction (BR) on pre-training performance for EquiformerV2, comparing results with and without knowledge distillation (KD). The table reports force MAE (meV/Å) across the OC20, OC22, ANI-1x, and Transition-1x datasets. Similar to GemNet-OC, we observe that later blocks contribute less to the final prediction, and KD mitigates performance degradation relative to the full model.}
    \centering
    \renewcommand{\arraystretch}{1.2} 
    \begin{tabular}{@{}lccccc@{}}
        \toprule
         & \# of Blocks & OC20 & OC22  & ANI-1x & Transition-1x \\ \midrule
        BR & 4 & 77.0 (\textcolor{negative}{-50.9}) & 71.5 (\textcolor{negative}{-40.4}) & 185.5 (\textcolor{negative}{-142.4}) & 69.6 (\textcolor{negative}{-52.0}) \\
        \hspace{1em}+ KD & 4 & 32.0 (\textcolor{positive}{-5.9}) & 41.4 (\textcolor{positive}{-10.3}) & 68.8 (\textcolor{positive}{-25.7}) & 26.9 (\textcolor{positive}{-9.3}) \\
        \midrule
        BR & 5 & 51.1 (\textcolor{negative}{-25.0}) & 54.1 (\textcolor{negative}{-23.0}) & 85.4 (\textcolor{negative}{-42.3}) & 40.8 (\textcolor{negative}{-23.2}) \\
        \hspace{1em}+ KD & 5 & 30.3 (\textcolor{positive}{-4.2}) & 38.2 (\textcolor{positive}{-7.1}) & 64.2 (\textcolor{positive}{-21.1}) & 24.3 (\textcolor{positive}{-6.7}) \\
        \midrule
        BR & 6 & 35.2 (\textcolor{negative}{-9.1}) & 40.9 (\textcolor{negative}{-9.8}) & 60.1 (\textcolor{negative}{-17.0}) & 25.6 (\textcolor{negative}{-8.0}) \\
        \hspace{1em}+ KD & 6 & 29.2 (\textcolor{positive}{-3.1}) & 35.3 (\textcolor{positive}{-4.2}) & 58.0 (\textcolor{positive}{-14.9}) & 21.8 (\textcolor{positive}{-4.2}) \\
        \midrule
        BR & 7 & 27.9 (\textcolor{negative}{-1.8}) & 33.4 (\textcolor{negative}{-2.3}) & 42.0 (\textcolor{positive}{1.1}) & 19.9 (\textcolor{negative}{-2.3}) \\
        \hspace{1em}+ KD & 7 & 27.4 (\textcolor{positive}{-1.3}) & 32.8 (\textcolor{positive}{-1.7}) & 53.4 (\textcolor{negative}{-10.3}) & 19.7 (\textcolor{positive}{-2.1}) \\
        \midrule
        EquiformerV2 (Teacher) & 8 & 26.1  & 31.1  & 43.1  & 17.6  \\
        \bottomrule
    \end{tabular}    
    \label{tab:pretraining_metrics_EqV2}
\end{table*}

\subsection{Evaluation on Downstream Tasks using EquiformerV2}
We evaluate the impact of our block reduction and pre-training knowledge distillation on downstream tasks, following the fine-tuning setup in Section~\ref{sec:main_results}. Distillation was applied only during pre-training to retain useful representations before fine-tuning.

\begin{figure*}[!h]
    \centering
    \vspace{0.4cm}
    \includegraphics[width=\textwidth]{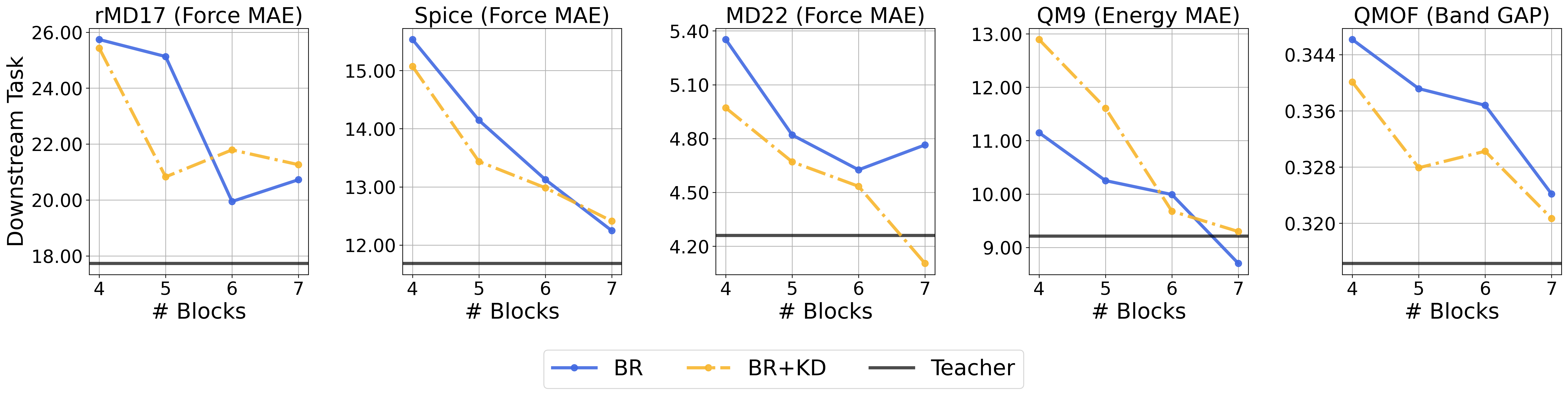}
    \caption{\textbf{Evaluation on downstream tasks using EquiformerV2.} We assess the performance of EquiformerV2 across various downstream tasks after applying block reduction (BR) and knowledge distillation (KD). Performance is reported in meV/Å for force targets, meV for the QM9 energy target, and eV for the QMOF band gap target. The teacher model refers to the original EquiformerV2 with 8 transformer blocks.}
    \label{fig:main_results_EqV2}
    \vspace{-0.2cm}
\end{figure*}

Appendix Figure~\ref{fig:main_results_EqV2} presents the results across multiple tasks. In Spice and QM9, the 7-block models achieve comparable or better performance than the 8-block teacher model, demonstrating that EquiformerV2 can be effectively compressed while maintaining accuracy for some tasks. In contrast, rMD17 and QMOF show a decline in performance as blocks are removed, even when distillation is applied. This suggests that the 31M-parameter teacher model was already compact, leaving little room for pruning without sacrificing accuracy. 

Distillation is most beneficial at 4 and 5 blocks, considerably recovering performance lost due to pruning. However, for 6 and 7 blocks, the gains from distillation diminish, as block reduction alone is sufficient for maintaining accuracy—except in MD22, where distillation leads to improved results over the full model.

These results highlight that while block reduction and knowledge distillation were effective for GemNet-OC, their impact on EquiformerV2 is mixed. One possible reason is that our pre-training duration for EquiformerV2 was limited to 8M samples compared to 120M samples in JMP, constrained by computational resources, which may have prevented the models from fully capturing useful representations. Additionally, EquiformerV2-31M might be too small to benefit from block reduction, as reducing blocks further constrains its capacity.

\begin{figure*}[!t]   

    \centering
    \includegraphics[width=\textwidth]{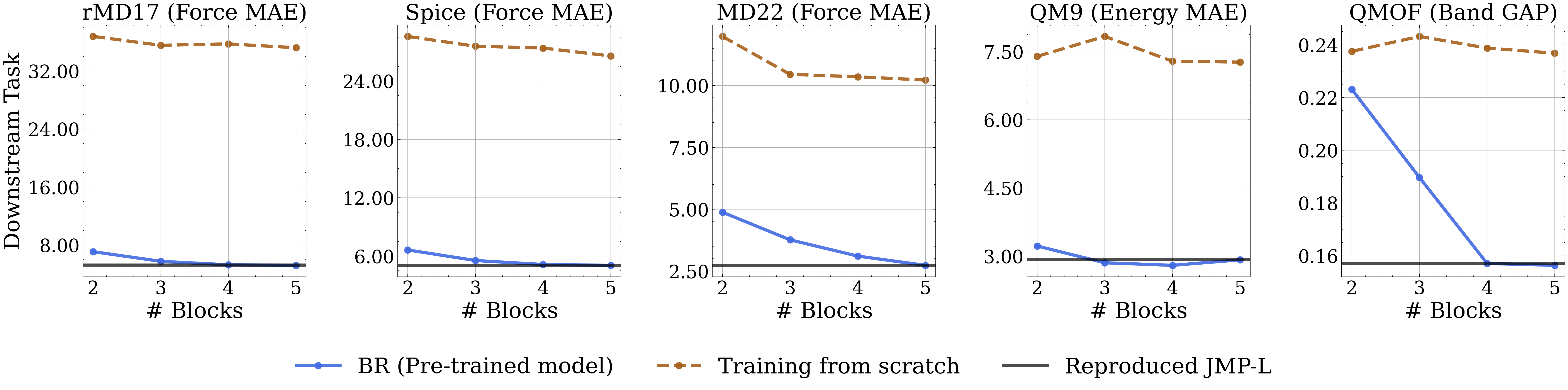}
    \vspace{-0.5cm}
    \caption{\textbf{Training from scratch vs. block reduction of a pre-trained model.} 
    Validation performance of smaller models across five downstream tasks when initialized from scratch (brown dashed line) versus when derived through block reduction from a pre-trained 6-block JMP-L model (blue line). The results highlight the importance of pre-training, as reduced models retain significantly better performance when initialized from a larger pre-trained backbone.
    } 
    
      \label{fig:scratch_analysis}
\end{figure*}

\begin{figure*}[!h]
    \centering
    \begin{minipage}[t]{0.48\textwidth}
        \centering
        \includegraphics[width=\linewidth]{figures/gemnet_gradient_analysis.png}
        \vspace{-0.5cm}
        \caption{\textbf{Block Relevance Analysis (Shown in Main Paper).} GradCAM-based importance scores for each block in JMP-L, previously presented in the main paper. The first block, $f_1$, represents the embedding output, while $f_2$ to $f_7$ correspond to the six interaction blocks of GemNet-OC.}

        \label{fig:block_imp_appendix}
    \end{minipage}
    \hfill
    \begin{minipage}[t]{0.48\textwidth}
        \centering
        \includegraphics[width=\linewidth]{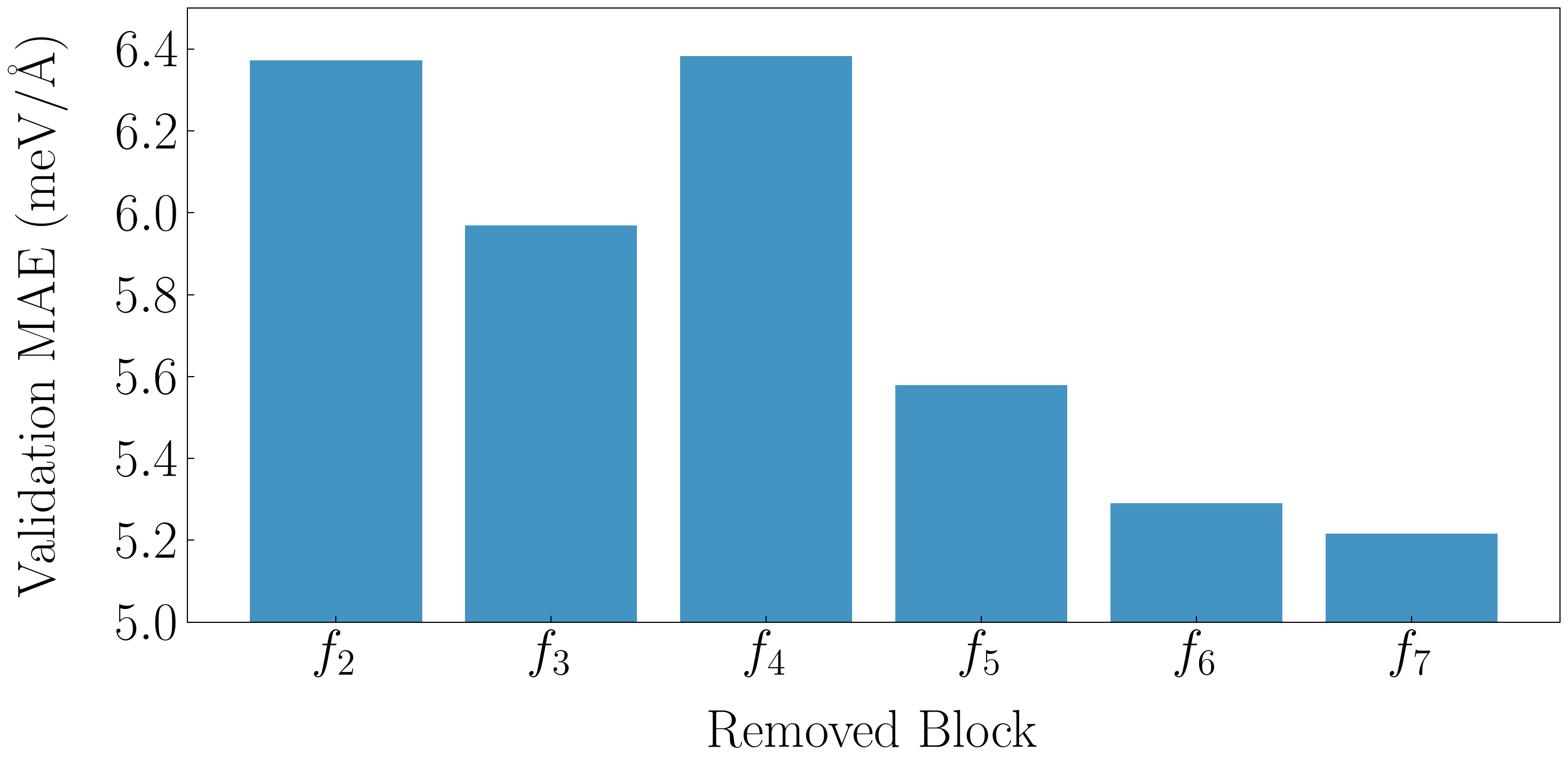}
        \vspace{-0.5cm}
        \caption{\textbf{Layer-Wise Ablation Study.} Validation MAE on rMD17 when individually removing each interaction block ($f_2$ to $f_7$) from the 6-block pre-trained GemNet-OC model. The results show that removing later blocks leads to smaller performance loss, supporting their lower importance as suggested by GradCAM.}
        \label{fig:layerwise_ablation}
    \end{minipage}
    \vspace{-0.2cm}
\end{figure*}

\section{Training from Scratch}
In the main paper, all experiments use models initialized from the JMP-L pre-trained checkpoint. Specifically, we apply block reduction to create smaller backbones by pruning layers from the full model—for example, constructing a 5-block GemNet-OC by removing the final block from a 6-block pre-trained backbone. But what if we instead train the 5-block model from scratch? How does this compare to pruning a pre-trained full-sized model?

Appendix Figure~\ref{fig:scratch_analysis} compares the downstream performance of models trained from scratch (brown dashed line) with those obtained via block reduction from a pre-trained model (blue line). Across all tasks, training from scratch consistently underperforms the block-reduced counterparts, often by a large margin. These results highlight the effectiveness of our block reduction strategy and demonstrate the importance of leveraging pre-trained weights, even when deploying smaller backbones.

\section{Layer-by-Layer Analysis of Block Reduction}

As discussed in Section~\ref{sec:interactionblock} of the main paper, our block reduction strategy is guided by block importance scores obtained using GradCAM. For instance, to create a 5-block GemNet-OC backbone, we remove the least important block—identified as the final block ($f_7$). However, this approach assumes that GradCAM accurately reflects functional relevance. Yet, removing a different block might lead to comparable downstream performance, which would cast doubt on the reliability of GradCAM for guiding pruning decisions. 

To evaluate the validity of the GradCAM scores, we conduct a layer-by-layer ablation, removing one interaction block at a time—starting from the first non-embedding layer ($f_2$) through to the final interaction block ($f_7$)—resulting in six separate runs. We note that removing $f_7$ matches the 5-block configuration used in the main paper. To manage computational cost, we restrict this analysis to the rMD17 dataset and fix the backbone size to 5 blocks.

In Appendix Figures~\ref{fig:block_imp_appendix} and~\ref{fig:layerwise_ablation} we present the block relevance scores and the corresponding downstream performance from the layer-by-layer ablation analysis, respectively. As predicted by GradCAM, removing later blocks leads to less performance degradation compared to earlier ones. Overall, the trend in performance drop from the first interaction block ($f_2$) to the last ($f_7$) aligns well with the relevance scores—except for $f_3$ and $f_4$, where removing $f_4$ leads to a larger drop than $f_3$. This suggests that GradCAM provides more reliable importance estimates for deeper blocks closer to the prediction head.

\end{document}